\documentclass{llncs}
\usepackage{graphicx}

\usepackage{tikz}
\usepackage{comment}
\usepackage{amsmath,amssymb} 
\usepackage{color}
\usepackage{bm}
\usepackage{booktabs}
\usepackage{multirow}
\usepackage{makecell}
\usepackage{wrapfig}
\usepackage{bbm}
\usepackage{bbding}
\usepackage{threeparttable}
\usepackage{xspace}
\usepackage{wrapfig}
\usepackage{setspace}
\usepackage[accsupp]{axessibility}  

\begin{document}
\pagestyle{headings}
\mainmatter

\title{Attention-based Class Activation Diffusion for Weakly-Supervised Semantic Segmentation}


\titlerunning{Attention-based Class Activation Diffusion for Weakly-Supervised Semantic Segmentation} 
\authorrunning{J. Huang et al.}
\author{Jianqiang Huang, Jian Wang, Qianru Sun and Hanwang Zhang}
\institute{Nanyang Technological University, Singapore
\\
\email{jianqiang.jqh@gmail.com, hanwangzhang@ntu.edu.sg}
\\
\and DAMO Academy, Alibaba Group, China
\\
\email{xuefei.wj@alibaba-inc.com}
\\
\and Singapore Management University, Singapore
\\
\email{qianrusun@smu.edu.sg}}

\maketitle

\begin{abstract}
Extracting class activation maps (CAM)~\cite{cam} is a key step for weakly-supervised semantic segmentation (WSSS). The CAM of convolution neural networks fails to capture long-range feature dependency on the image and result in the coverage on only foreground object parts, i.e., a lot of false negatives.
An intuitive solution is ``coupling'' the CAM with the long-range attention matrix of visual transformers (ViT)~\cite{vit}.
%
%
We find that the direct ``coupling'', e.g., pixel-wise multiplication of attention and activation~\cite{ts-cam}, achieves a more global coverage (on the foreground), but unfortunately goes with a great increase of false positives, i.e., background pixels are mistakenly included.
%
This paper aims to tackle this issue.
%
It proposes a new method to couple \textbf{CAM} and \textbf{A}ttention matrix in a probabilistic \textbf{D}iffusion way, and dub it \textbf{AD-CAM}.
Intuitively, it integrates \underline{ViT attention} and \underline{CAM activation} in a \emph{conservative} and \emph{convincing} way.
\emph{Conservative} is achieved by refining the attention between a pair of pixels based on their respective attentions to common neighbors, where the intuition is two pixels having very different neighborhoods are rarely dependent, i.e., their attention should be reduced.
\emph{Convincing} is achieved by diffusing a pixel's activation to its neighbors (on the CAM) in proportion to the corresponding attentions (on the AM).
In experiments, our results on two challenging WSSS benchmarks PASCAL VOC and MS~COCO show that AD-CAM as pseudo labels can yield stronger WSSS models than the state-of-the-art variants of CAM.
\end{abstract}
\section{Introduction}
\label{sec_intro}


Semantic segmentation aims to infer the object label on each pixel of an image.
Its fully-supervised labels are expensive to obtain.
A lot of recent research is thus focused on the training settings with only coarse or weak labels such as bounding boxes~\cite{bbox1,bbox2}, scribbles~\cite{scribble1,scribble2} and image-level class labels~\cite{sec,irn,auxiliary,complementary,group,matters,recam}.
This paper aims to tackle the last setting which is the most challenging one, but calls the task simply as weakly-supervised semantic segmentation (WSSS).
%
The prevailing pipeline of WSSS using image labels includes a few steps: 1)~loading a large-scale classification model pre-trained e.g. on ImageNet~\cite{imagenet}; 
2)~fine-tuning the model on the WSSS dataset where each sample has an image and a multi-hot label that represents co-occurring classes on the image;
%
3)~extracting a class activation map (CAM)~\cite{cam} for each occurring class, and then hard-thresholding it to a 0-1 mask (note that it is optional to refine the mask via erosion and expansion techniques~\cite{irn,anti}); 
and 4) taking such masks as pseudo labels to train the semantic segmentation model in a standard fully-supervised pipeline~\cite{deeplabv2,deeplabv3p,upernet}. 
The quality of the segmentation model thus depends on two major factors: the multi-label classification model and the extraction method of CAM.
In this work, we inspect the problems in existing CNN- and ViT-based classification models when they are used to extract CAM.

\begin{figure*}[!t]
    \centering
    \includegraphics[width=1.0\linewidth]{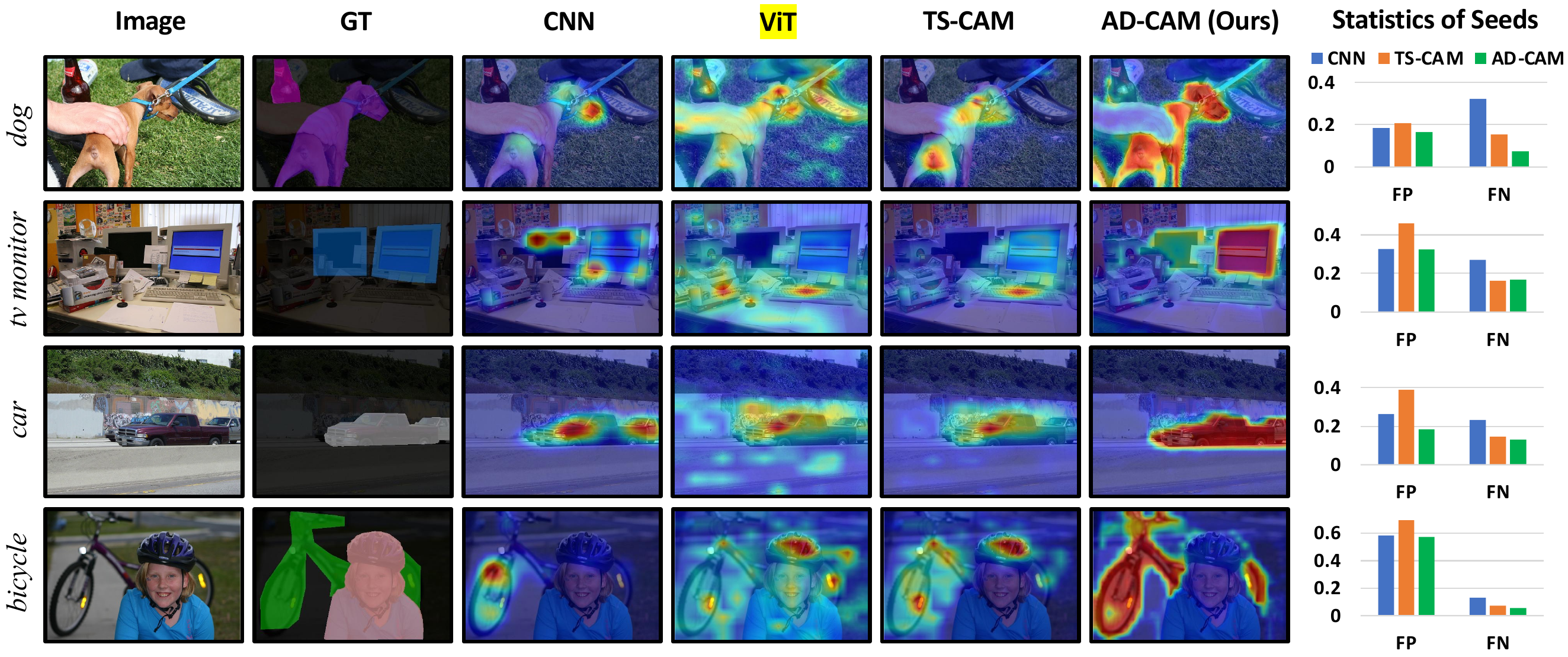}
    \vspace{-4mm}
    \caption{CAM heatmaps (i.e., seeds of WSSS) generated by different methods: 1) CNN column is by ResNet-50~\cite{cam}; 2) ViT column shows the model attention from visual transformer~\cite{vit} \underline{without distinguishing classes}; 3) TS-CAM column combines attention and activation on a cascaded CNN-ViT architecture~\cite{ts-cam} which is a close related work to ours; and 4) our AD-CAM column shows our method results based on a double-branch CNN-ViT architecture~\cite{conformer}. 
    For each individual class, we display the statistics of false positive (FP) and false negative (FN) pixels on 0-1 seed masks (generated by applying the optimal threshold of each method on the heatmaps). In experiments, we also compare TS-CAM and AD-CAM on the same network architecture.
    }
    \label{fig:teaser}
    \vspace{-4mm}
\end{figure*}

\textbf{Problems.}
Figure~\ref{fig:teaser} demonstrates example CAM heatmaps generated by using ResNet-50, ViT~\cite{vit}, TS-CAM~\cite{ts-cam} on a cascaded CNN-ViT architecture, and our AD-CAM on a two-branch CNN-ViT architecture~\cite{conformer}, respectively.
For each individual class, we also show the statistics of the false positive (FP) and false negative (FN) pixels on 0-1 masks (generated by applying the optimal hard threshold of each method on the heatmaps).
We have several observations from Figure~\ref{fig:teaser}. 
1)~CNN has local activations and its 0-1 mask suffers from a great number of FN pixels, e.g., its activations on \emph{dog} are mainly on the \emph{dog head} and its FP rate is significantly higher than others.
%
%
2)~ViT based on long-range attention\footnote{
Please note that the activation map by ViT models is generated by calculating the attention matrix between classification and all tokens and then upsampling the attention matrix ($14\times14$ in our case) to the image size ($H\times W$).
} sheds light on activating more regions on the image (than CNN). However, its effectiveness of reducing FN goes along with the increase of FP and it does not distinguish classes.
%
%
3)~TS-CAM, using a basic operation (i.e., dot product) to
couple the activation maps of CNN and ViT, does not solve the key problem. Thanks to the coupling with CNN, TS-CAM can distinguish classes. Yet, it still has to trade off FN and FP. For example, its FP rates increase greatly (over CNN) for \emph{tv monitor}, \emph{car} and \emph{bicycle}, and slightly for \emph{dog}, when it reduces FN\footnote{Please note that our demonstration results in Figure~\ref{fig:teaser} are all based on the optimal thresholds that are carefully chosen. We leave the implementation details in supplementary materials.}.
The visual examples show TS-CAM has a lower FP than ViT.

\textbf{Motivations.}
From Figure~\ref{fig:teaser}, we further inspect the activated pixels of ViT that contain all semantics on the image without distinguishing classes.
We find a large number of long-range but spurious correlations either between labeled objects such as \emph{dog} and \emph{tv monitor} and non-labeled objects such as the \emph{slipper} beside the \emph{dog} and the \emph{keyboard} beside \emph{tv monitor}, or between co-occuring labeled objects such as \emph{person} (especially the \emph{helmet}) and \emph{bicycle} in the bottom example. 
In many cases, they even dominate the coupling results if the coupling operation is straight-forward. For example, after TS-CAM, the activation on the \emph{tv monitor} image is almost overtaken by the \emph{keyboard}, and the \emph{bicycle} case is overtaken by the \emph{helmet}.
%
%
%
%
We think the crux behind has two aspects. 1) ViT classifiers tend to use every cue of the object image, no matter how far it is, to make the prediction even through some of cues are from the context. 
2) Using basic coupling operations, e.g., dot product, can not prevent the spurious and dominating cues from merging into CAM.
%
%

\textbf{Solutions.}
To solve these issues, we introduce 
two novel modules in our AD-CAM method, accordingly.
First, we propose a new measurement on the confidence of ViT attention, and call it co-neighbor similarity. It indicates the attention between two pixels is confident only when their respective neighborhoods are similar, i.e., sharing a great number of nearest neighbors. 
This makes AD-CAM (when using this attention) more 
conservative and prudent against FP attentions.
Second, we propose a more convincing operation (than dot product) to couple the attention of ViT to the CAM of CNN.
We diffuse each pixel's class activation value to its nearest neighbors on the CAM, in proportional to the corresponding values on the attention matrix.
Intuitively, each pixel computes its diffusion probabilities to nearest neighbors based on attentions. Each probability determines how much activation to diffuse from the pixel to this neighbor.
After $n$ steps of diffusion, where $n$ is a hyperparameter, we obtain the final class activation maps of AD-CAM\footnote{AD-CAM is the name of method as well as the name of the produced class activation maps using the method.}. Then, we proceed the following steps of WSSS: refining seed masks; training semantic segmentation models; and evaluating them.

In summary, our technical contributions are two folds. 
1) A new co-neighbor similarity to measure the confidence of attention produced by ViT. This helps to alleviate false positive issues. 
2) A new activation diffusion operation to incorporate the attention of ViT to expand the CAM of CNN (note that here the attention indicates the refined version by using co-neighbor similarity). It solves the issue of false negative in CAM and maintains a low false positive rate when incorporating long-range attentions. It yields more convincing results than using doc product.

\section{Related Work}
\label{sec_related}

The training of multi-label classification and semantic segmentation models are almost uniform in the related works of WSSS~\cite{conta,anti,rib}.
Below, we focus on the introduction of their particular steps of generating or refining CAM.

\noindent
\textbf{CAM Generation and Refinement.}
Vanilla CAM~\cite{cam} is the pioneer work of generating the activation map for a specific class. It uses the weight of the last FC layer (i.e., the classifier in an end-to-end CNN model) to average the feature maps and then generate a mask through spatial-wise normalization and hard thresholding (see Section~\ref{sec_preli}). 
%
Such masks are usually taken as seeds, and a lot of follow-up works are focused on seed refinement. 
GAIN~\cite{gain} applies the CAM on original images to generate masked images, and minimizes the model prediction scores on them, forcing the model to capture the features in other regions (outside the current CAM) in the new training. A similar idea was used in erasing-based methods~\cite{adv_erasing,cse,selferasing,adv_erasing2}. Score-CAM~\cite{scorecam} proposed to replace the FC weights used in vanilla CAM with a new set of scores predicted from the images masked by channel-wise (not class-specific) activation maps. EDAM~\cite{edam} and ReCAM~\cite{recam} proposed to improve the CAM by adding an extra classifier. One category of refinement methods~\cite{psa,irn,auxiliary,bes} propagate the object regions in the seed to semantically similar pixels in the neighborhood. It is achieved by the random walk~\cite{randomwalk} on a transition matrix where each element is an affinity score. The related methods have different designs of this matrix. PSA~\cite{psa} is an AffinityNet to predict semantic affinities between adjacent pixels. 
IRN~\cite{irn} is an inter-pixel relation network to estimate class boundary maps based on which it computes affinities. Another method is BES~\cite{bes} that learns to predict boundary maps by using CAM as pseudo ground truth. Our AD-CAM also apply a diffusion strategy, but we use the distilled attention matrix to capture the pixel-pixel affinity without introducing any additional networks. The CAM generated by CAM can be further refined by existing refinement methods. Another category of refinement methods~\cite{nsrom,dsrg,attenbn,ficklenet,splitting,suppression} utilize saliency maps~\cite{saliency1,saliency2}. EPS~\cite{eps} proposed a joint training strategy to combine CAM and saliency maps. EDAM~\cite{edam} introduced a post-processing method to integrate the confident areas in the saliency map into CAM. A more recent category of methods leverage iterative post-processing to refine CAM. OOA~\cite{ooa} ensembles the CAM generated in multiple training iterations. CONTA~\cite{conta} iterated through the whole process of WSSS including a sequence of model training and inference. AdvCAM~\cite{anti} used the gradients with respect to the input image to perturb the image, and iteratively find newly activated pixels.

\noindent
\textbf{CAM Generation based on Attention (of ViT).}
It is a known problem that using CNN to extract CAM fails to mask the complete object regions, i.e., suffers from a great number of false negative pixels in the results.
ViT is different with CNN and can capture long-range dependencies on the input image~\cite{vit}. 
A pioneer method of combining the self-attention of ViT with CNN (to extract CAM) is called TS-CAM~\cite{ts-cam}.
%
Its key operation is multiplying CAM by the class-wise attention maps (extracted by using class tokens). This naive pixel-wise operation however results a lot of false positive in the results.
Our AD-CAM also uses the attention of ViT to improve CAM. We highlight two differences with TS-CAM. 
1) We propose a new method of using the direct output of ViT, i.e., self-attention matrix, in AD-CAM. While TS-CAM needs to compute an additional ``attention activation map'' using class token and image patch token. Therefore, ours is more generic to different ViT models.
2) Our AD-CAM aims for tackling WSSS tasks. While TS-CAM was for weakly-supervised object localization (WSOL) tasks which use single object images and the result models can not be adapted for WSSS (because the class token in multi-label ViT do not distinguish co-occurring classes on the image).
A more recent work GETAM~\cite{getam} is the first ViT-based approach for WSSS. It improves CAM with better object shapes by coupling the ``attention activation map'' and its class-wise gradient maps. 
Its method of using self-attention is different with ours. We compare to it in experiments.
\section{Preliminaries: A Pipeline of Training WSSS Models}
\label{sec_preli}

\noindent
\textbf{Seed Generation.}
Extracting CAM~\cite{cam} is based on a trained multi-label classification model. The first step is thus to train this model.
It is an end-to-end training on the network consisted of a feature extractor (backbone), 
a global average pooling (GAP) layer, and a class prediction layer (e.g., an FC layer on ResNet-50~\cite{resnet}).
%
The loss used for training is usually a binary cross-entropy (BCE) loss:
{\small
\begin{equation}
    \mathcal{L}_{bce}=-\frac{1}{K} \sum_{k=1}^{K} y \left[ k \right] \log \sigma\left(\hat{y}[k]\right)+\left(1-y[k]\right) \log \left[1-\sigma\left(\hat{y}[k]\right)\right],
\label{eq:bce}
\end{equation}}

\noindent
where \(\hat{y}[k]\) denotes the prediction logit for the \(k\)-th class, \(\sigma(\cdot)\) is the Sigmoid function, and \(K\) represents the total number of object classes on WSSS dataset. 
\(y[k]\in\{0,1\}\) is the image-level label for the \(k\)-th object class, where 1 denotes the class is present in the image and 0 otherwise.

When the model is trained, each training image $\bm{x}$ is again fed into it to extract CAM for the $k$-th class:
\begin{equation} \label{equation:cam}
    \operatorname{CAM}_k(\bm{x})=\frac{\operatorname{ReLU}\left(\bm{F}_k\right)}{\max \left(\operatorname{ReLU}\left(\bm{F}_k\right)\right)}, \bm{F}_k=\mathbf{w}_{k}^{\top}f(\bm{x}),
\end{equation}
where \(\mathbf{w}_{k}\) denotes the classification weights of the \(k\)-th class, and \(f(\bm{x})\) denotes the feature maps before GAP.
Please note that, for simplicity, we assume the classification head of the model is always a single FC layer, and use \(\mathbf{w}\) to denote its learned weights.

\noindent
\textbf{Pseudo Mask Generation.} 
Taking CAM as seed, there are mainly two options to produce pseudo masks: 1) thresholding CAM to be 0-1 masks, and 2) refining CAM with IRNet~\cite{irn} and then thresholding the refined CAM to be 0-1 masks.

\noindent
\textbf{Semantic Segmentation.} 
This is the last step of WSSS. The generated pseudo masks are used to train a standard semantic segmentation model, e.g., any DeepLab variants~\cite{deeplabv2,deeplabv3p}, in a fully-supervised fashion. The objective function is:
{\small
\begin{equation}
    \mathcal{L}_{ss}=-\frac{1}{hw} \sum_{i=1}^{h} \sum_{j=1}^{w} \frac{1}{K\!+\!1} \sum_{k=1}^{K+1} y_{i,j}[k]  \log \frac{\exp (\hat{y}_{i,j}[k])}{\sum_{k} \exp (\hat{y}_{i,j}[k])},
\end{equation}}

\noindent
where $y_{i,j}$ and $\hat{y}_{i,j}$ denote the label and the prediction logit at pixel $(i,j)$, respectively. $y_{i,j}[k]$ and $\hat{y}_{i,j}[k]$ denote the $k$-th element of $y_{i,j}$ and $\hat{y}_{i,j}$, respectively. $h$ and $w$ are the height and width of the image. $K$ is the total number of classes in WSSS dataset. $K\!+\!1$ means taking the \texttt{background} as one more ``class''.


\begin{figure*}[!t]
    \centering
    \includegraphics[width=1.0\linewidth]{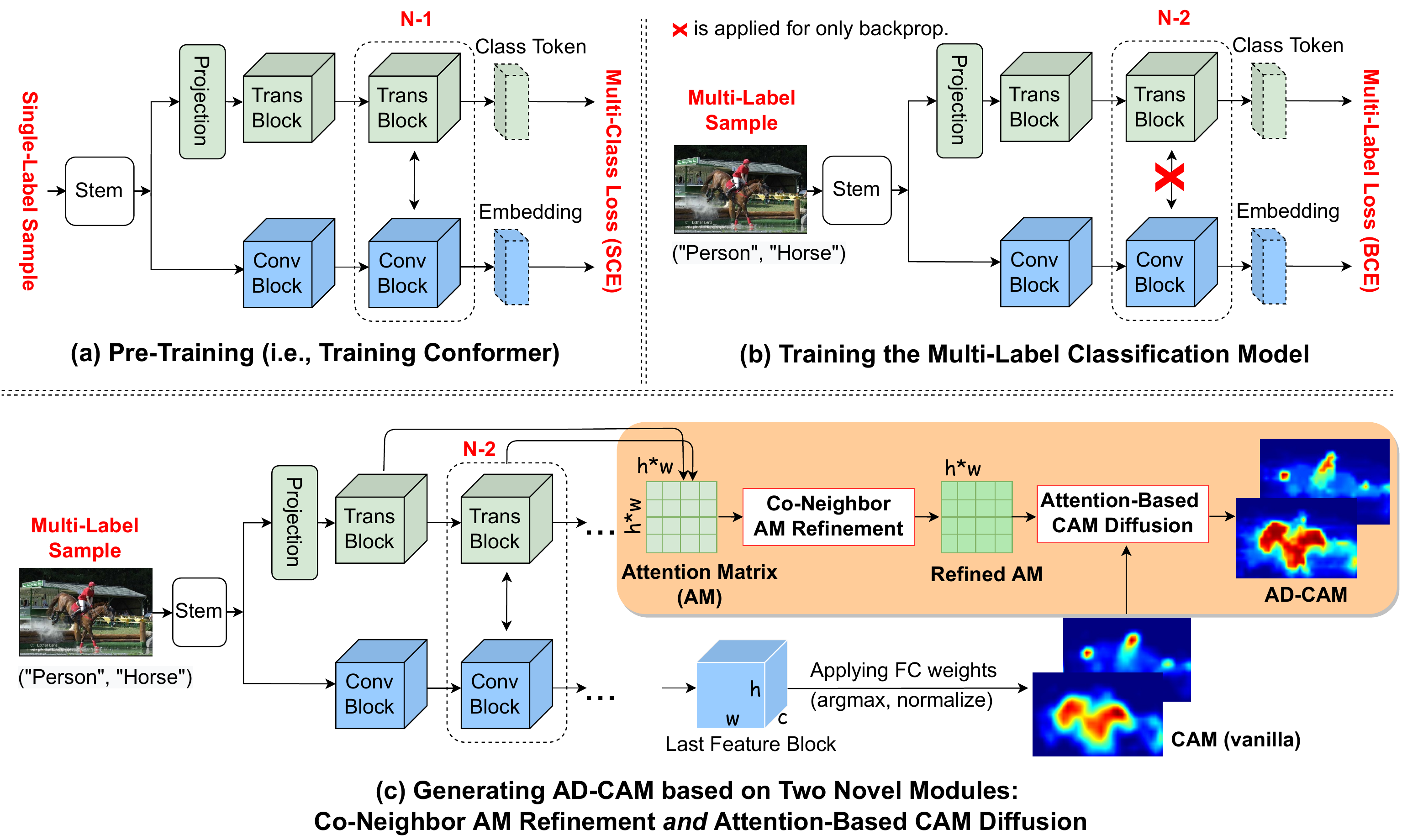}
    \vspace{-4mm}
    \caption{Our training pipeline is consisted of three steps. (a) Pre-training the backbone of two branches: one for CNN and one for ViT. We directly load the trained weights published in \cite{conformer}. Please note we apply this backbone also for related works to keep a fair comparison. 
    (b) Training a multi-label classification model on the WSSS datasets such as MS~COCO~\cite{mscoco}. 
    (c) Extracting AD-CAM, i.e., seed masks for WSSS. We highlight our proposed modules in the orange box. ``AM'' stands for attention matrix.}
    \label{fig:overview}
\end{figure*}

\section{AD-CAM}
The overview of our approach is shown in Figure~\ref{fig:overview}.
It does not include the steps of seed refinement and semantic segmentation model training as they are standard as in related works~\cite{recam,anti-advcam}.
The architecture of AD-CAM is based on CNN and ViT, and is called Conformer~\cite{conformer}. The procedure to deriving AD-CAM has three steps. 
Step 1: Pre-training the backbone of Conformer using a standard multi-class classification pipeline on ImageNet~\cite{imagenet}. 
To save the computational cost, we directly load the trained weights published in \cite{conformer}. 
Please note we apply this stronger backbone also for related works, such as CAM~\cite{cam} and TS-CAM~\cite{ts-cam}, to keep a fair comparison. 
Step 2: Training a multi-label classification model on WSSS datasets, as formulated in Section~\ref{sec_preli}, based on the pre-trained Conformer.
Step 3: Extracting AD-CAM for all training images. In Figure~\ref{fig:overview}~(c), we highlight our two new modules in the orange box. ``AM'' stands for the attention matrix of ViT.
In the following, we elaborate Step 2 and Step 3 in three subsections.

\subsection{Training Multi-Label Classifiers}


\noindent
\textbf{Feature Extraction.}
Given an input image $\bm{x}$ and its multi-hot class label $\bm{y}\in\{0,1\}^{K}$, $K$ is the total number of foreground classes in the dataset. We denote the output of CNN feature encoder as $f_{\text{CNN}}(\bm{x}) \in \mathbb{R}^{h \times w \times C}$ and the classification token of ViT as $f_{\text{ViT}}(\bm{x}) \in \mathbb{R}^{1 \times C}$. $C$ denotes the number of channels. $h$ and $w$ are height and width of feature map, respectively.  

\noindent
\textbf{Prediction.} 
The prediction on the CNN branch is formulated as:
\begin{equation}
    \bm{\hat{y}}_{\text{CNN}} = \operatorname{FC_1}(\operatorname{GAP}(f_{\text{CNN}}(\bm{x}))),
\end{equation}
The prediction on the ViT branch is formulated as:
\begin{equation}
    \bm{\hat{y}}_{\text{ViT}} = \operatorname{FC_2}(f_{\text{ViT}}(\bm{x})).
\end{equation}
where $\text{FC}_1$ and $\text{FC}_2$ denote classification heads.

\noindent
\textbf{Optimization.} We use these predictions to compute the binary cross entropy (BCE) losses, as in Eq.~\eqref{eq:bce}, for two branches respectively.
We highlight that when backpropagating gradients, we detach the interaction between two branches, as shown by a red cross in Figure~\ref{fig:overview}~(b). Our aim is to enlarge the ``difference'' between two branches that may compensate each other
in the final coupling step. We evaluate this in Section~\ref{sec_ablation}.

\subsection{Extracting CAM and AM}

As shown in Figure~\ref{fig:overview}~(c), we need to extract CAM from CNN branch and AM from ViT branch, before the orange box.

\noindent
\textbf{CAM.}
This is similar to the case in an independent convolutional backbone. Following Eq.~\eqref{equation:cam},
we extract $\operatorname{CAM}_k(\bm{x})\in \mathbb{R}^{h \times w}$ for each class $k$ given the feature $f_{\text{CNN}}(\bm{x})$ and the 
classifier weights $\mathbf{w}_k$ on the CNN branch.
%
For brevity, we denote the $\operatorname{CAM}_k(\bm{x})$ as $\bm{M}_k$ omitting the notation of input image. 

\noindent
\textbf{AM.}
For the $i$-th transformer module, we use $\bm{A}_{i}\in \mathbb{R}^{(hw)\times (hw)}$ to represent the 
average results over its multi-head attentions.
%
Each value on this matrix denotes the correlation between two feature pixels (or more precisely, two image patches).
We aggregate the AMs of all transformer modules as $\bm{A} = \frac{1}{L}\sum_{i=1}^{L}\bm{A}_{i}$, 
where $L$ is the total number of transformer modules on the ViT branch. 


\subsection{Generating AD-CAM}
This step is illustrated in the orange box of Figure~\ref{fig:overview}~(c). We have two novel modules to generate AD-CAM. 

\noindent
\textbf{Co-Neighbor AM Refinement.} 
As inspected in Figure~\ref{fig:teaser} and mentioned in Section~\ref{sec_intro}, the AM often contain false dependencies, and these dependencies may corrupt vanilla CAM especially when the coupling method (of AM and CAM) is improper.
To alleviate the false dependencies, we introduce a new measurement on the confidence of attention, called co-neighbor similarity.
It indicates the attention between two pixels is confident only when their respective neighborhoods are similar, i.e., sharing a great number of nearest neighbors. In the following, we first elaborate the definition of this similarity and then introduce how to use it to refine AM.

For a pair of feature pixels $i$ and $j$, their co-neighbor similarity is defined as:
\begin{equation}
    \bm{S}_{ij} = \sum_{k}\sqrt{\bm{A}_{ik}\bm{A}_{jk}},
\end{equation}
In $\bm{A}$, the sum on each row equals to 1. 
Thus, we have $\bm{S}_{ij} \in [0,1]$ by Cauchy-Schwarz inequality and $\bm{S}_{ij} = 1$ if and only if $\bm{A}_{ik} = \bm{A}_{jk}$ for any $k$.
\begin{theorem}
\label{thm:1}
If pixel $i$ and pixel $j$ has $n$ neighbors with equal attention $\frac{1}{n}$, then we have $$\bm{S}_{ij} = \frac{2\text{Jaccard}(i,j)}{(1+\text{Jaccard}(i,j))},$$
where $\text{Jaccard}(i,j)$ is the Jaccard similarity between $i$ and $j$.
\end{theorem}
By theorem \ref{thm:1}, we can see that $\bm{S}_{ij}$ is a generalization of Jaccard similarity, which implies that $\bm{S}_{ij}$ is a valid metric by considering neighboring information. 

\begin{figure*}[!t]
    \centering
    \includegraphics[width=1.0\linewidth]{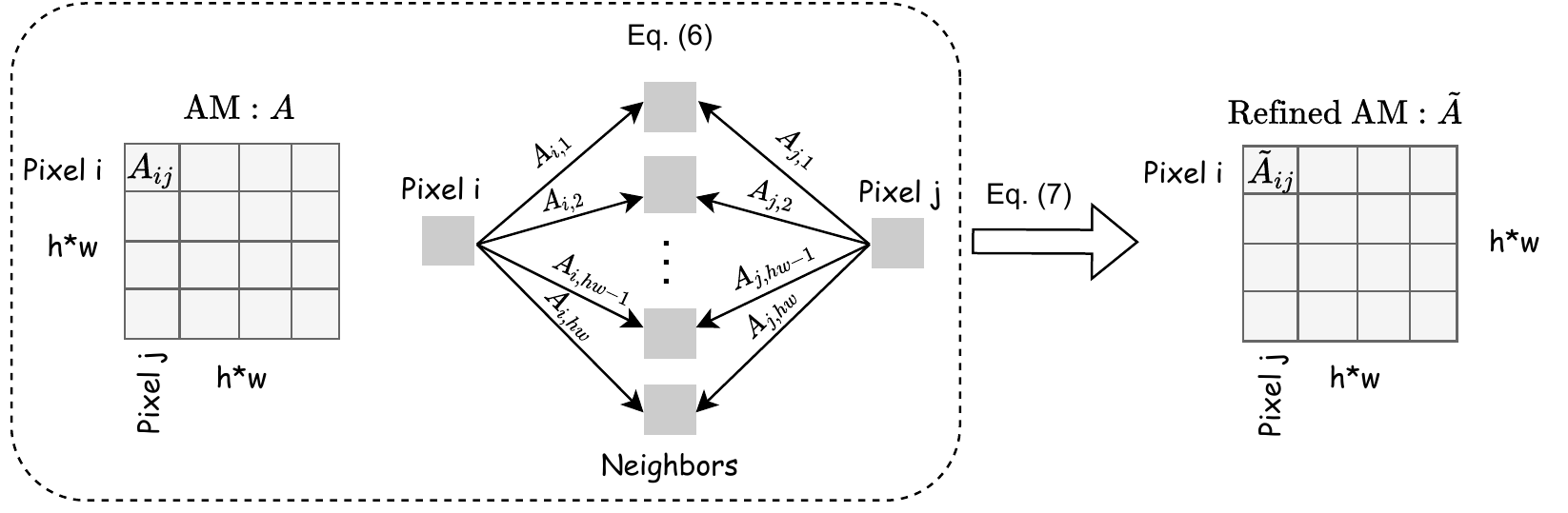}
    \vspace{-4mm}
    \caption{Illustrations of using our co-neighbor similarity to refine attention matrix (AM).}
    \label{fig:coneighbor}
\end{figure*}

Then, based on $\bm{S}$, we conduct the co-neighbor AM refinement as follows,
\begin{equation}
    \tilde{\bm{A}} = \text{NORM}(\text{TOP}(\bm{S},k)\otimes\bm{A}),
\end{equation}
where $\text{TOP}(\bm{S},k)$ is an operation to search the top-$k$ nearest neighbor pixels for each pixel, based on the similarities in $\bm{S}$. $\otimes$ denotes Hadamard product. 
$\text{NORM}$ is to make each row of $\tilde{\bm{A}}$ sum up to 1.
In $\tilde{\bm{A}}$, each value is refined to be more confident (than that in $\bm{A}$) based on its neighborhood distribution.

\noindent
\textbf{Attention-based CAM Diffusion.} 
Each row of $\tilde{\bm{A}}$ sum up to 1. We can build a diffusion process based on $\tilde{\bm{A}}$. Our aim is to diffuse the activation of each pixel to its neighbors (on $\bm{M}_k$) in proportional to the corresponding attentions (from this pixel to its neighbors in $\tilde{\bm{A}}$). Thus, we have
\begin{equation}
    \bm{M}_k(i)^{(t+1)} = \sum_{j}\tilde{\bm{A}}_{ji}\bm{M}_k(i)^{(t)},
\end{equation}
where $\bm{M}_k^{(0)}$ is the initial (vanilla) CAM of the $k$-th class. During diffusion, each pixel $i$ diffuses its own activation to neighbors and also receives activation from the pixels which take $i$ as nearest neighbors in $\tilde{\bm{A}}$.
After $T$ steps of diffusion, we obtain the AD-CAM $\bm{M}'_{k}$ where $k = 1,2,\cdots, K$ and $T$ is a hyperparameter.

\subsection{Refining AD-CAM (Optional)}
In recent works of WSSS, it is almost a standard step to leverage
IRNet~\cite{irn} to refine the mask before training any segmentation models.
In our case, we can take AD-CAM $\bm{M}'$ as the input and train an IRNet
to estimate the class boundary maps $\mathcal{B}$. 
We omit the training details of IRNet for brevity. 
Then as in~\cite{irn}, we apply a random walk to refine AD-CAM with $\mathcal{B}$ and a transition probability matrix $\mathbf{T}$:
\begin{equation}
    \label{eq:irn}
    \operatorname{vec}\left(\bm{\hat{M}}_k\right)=\mathbf{T}^{l} \cdot \operatorname{vec}\left(\bm{M}'_{k} \otimes(1-\mathcal{B})\right),
\end{equation}
where $l$ is the number of iterations and $\operatorname{vec}(\cdot)$ denotes vectorization.
Finally, we use $\{\bm{\hat{M}}_k\}$, where $k$ denotes every positive class in the image, as pseudo masks to train semantic segmentation models in a supervised way.

\section{Experiments}
\label{sec_exper}

\subsection{Datasets and Settings}
\label{sec_data_setting}

\noindent
\textbf{Datasets.} 
We use two commonly used datasets to evaluate our method: PASCAL VOC 2012~\cite{voc} and MS~COCO 2014~\cite{mscoco}. \textbf{VOC} contains $20$ foreground object classes and $1$ \texttt{background} class. It has $1,464$, $1,449$, and $1,456$ samples in \texttt{train}, \texttt{val}, and \texttt{test} sets, respectively. 
In our estimation, there are around $59.1\%$ single-label images, i.e., with one-hot labels, on this dataset.
Following our related works~\cite{irn,conta,anti,recam}, we exploit an extra training set with $10,582$ training images that are provided by Hariharen et al.~\cite{voc_aug}. \textbf{MS~COCO} has $80$ object classes and $1$ \texttt{background} class. Its \textit{train} and \textit{val} sets contains $80k$ and $40k$ samples, respectively. 
Compared to VOC, this dataset is more challenging: 1) it has more classes in total; and 2) there are only $20.3\%$ single-label images, i.e., with one-hot labels, and the average object number per image is $2.9$.
Please note that on both datasets, we use the samples with only image-level labels for training, i.e., the most challenging setting of WSSS.

\noindent
\textbf{Evaluation Metrics.} 
We have mainly two evaluation steps. \textit{Mask Generation:} We generate pseudo masks, including Seed Masks and final Pseudo Masks (also called Refined Masks by IRNet~\cite{irn}), for the images in the \texttt{train} set and use corresponding ground truth masks to compute mIoU. 
\textit{Semantic Segmentation:} We train the segmentation model using Pseudo Masks, use the resultant model to predict masks for the images in \texttt{val} or \texttt{test} sets, and compute mIoU based on ground truth masks. 
We provide the results of F1 and pixel accuracy in the supplementary materials.

\noindent
\textbf{Network Architectures.}
In the step of seed mask generation, we use Conformer~\cite{conformer} as backbone, and drop the last Transformer and Conv block
to make its output feature map size to be $\frac{1}{16}$ of image size. This is to match the feature resolution as in conventional CNN-based CAM variants~\cite{resnet}.
In the step of semantic segmentation, we employ ResNet-101 based DeepLabV2, pre-trained on ImageNet, following the common practice in related works~\cite{irn,adv_erasing2,anti-advcam}).

\noindent
\textbf{Implementation Details.}
%
%
As shown in Figure~\ref{fig:overview}~(b), we fine-tune the Conformer on multi-label classification tasks, e.g., on VOC and MS~COCO. 
In this step, we use AdamW optimizer with 1) a constant learning rate of 2e-4 on the ViT branch, and 2) a polynomial decay strategy from the initial learning rate of 5e-4 on the CNN branch.
It takes in total 5 epochs in this step (much shorter than in the pre-training step).
We use the same techniques of image pre-processing and augmentation technique as in IRN ~\cite{irn}.
The only difference is that we crop each input image 384$\times $384 pixels to adapt the pre-trained model. 

As shown in Figure~\ref{fig:overview}~(c), it is the step of generating AD-CAM. We have two key hyperparameters. 1) For the top-$k$ nearest neighbors in co-neighbor AM refinement, we set $k$ as $50$. 2) For the attention-based CAM diffusion, we set the number of diffusion steps $T$ as $2$.

Please note that we put the implementation details of training semantic segmentation models in the supplementary materials. 


\subsection{Results and Analyses}
\label{sec_result_analysis}

\begin{table}
  \centering
  \caption{Comparing our method with baselines in terms of seed, pseudo label (P.L.) and the final segmentation result mIoU (\%) on VOC dataset (the results of MS~COCO are in supplementary file). * denotes the backbone for segmentation is transformer. ** denotes the backbone of DeepLabV2 for VOC has been pre-trained on MS~COCO, which applies to ``Seg(\texttt{val})'' and ``Seg(\texttt{test})'' columns.}
  %
  \label{table_all_miou}
  \setlength{\tabcolsep}{1.5mm}{
  \renewcommand\arraystretch{1}
  \begin{tabular}{lcccc}
    \toprule
    Methods & \texttt{Seed} & \texttt{P.L.} & Seg(\texttt{val})& Seg(\texttt{test}) \\
    \midrule
    CAM(ResNet-50)~\cite{cam,resnet}          & 48.8  & 66.3  & 63.5  & 64.8     \\
    AdvCAM~\cite{anti}      & 55.6  & 69.9  & 68.1  & 68.0    \\
    RIB~\cite{rib}          & 56.5  & 70.6  & 68.3  & 68.6    \\
    ReCAM~\cite{recam}      & 54.8  & 70.9  & 68.7  & 68.5   \\
    \hline
    CAM(Conformer)~\cite{cam,conformer}         & 54.7  & 67.3  & 65.7     & 66.6          \\
    AD-CAM                  & 63.5  & 71.7  & 69.0  & 69.4       \\
    AD-CAM**                & 63.5     & 71.7     & 70.7  & 70.7 \\
    \bottomrule
  \end{tabular}
  }
\end{table}
\noindent
\textbf{Evaluating Seeds and Pseudo Masks.}
Table~\ref{table_all_miou} compares our initial seeds and pseudo labels with that obtained by state-of-the-art methods. Among them, AdvCAM~\cite{anti}, RIB~\cite{rib} and ReCAM~\cite{recam} are based on ResNet-50, TS-CAM~\cite{ts-cam} and GETAM~\cite{getam} are based on ViT. Both seeds and pseudo labels are generated from the training images of the PASCAL VOC and MS COCO datasets. For the inital seed, we report the best results by applying a range of thresholds to separate the foreground and background in the CAM $\{\bm{\hat{M}}_k\}$, as following to SEAM~\cite{seam}. 
It is clear to see that the initial seeds of our AD-CAM is much better than the CAM generated from the convolution head of Conformer, e.g., 8.4\% higher on VOC, which implies that our attention diffusion procedure is essential. And AD-CAM also outperforms a wide range of state-of-the-art methods. For all of the methods, the pseudo labels (P.L.) are obtained by a general process of IRN~\cite{irn}. Without doubts, AD-CAM achieves the best performance.

\noindent
\textbf{Evaluating WSSS Models on VOC.}
In the middle columns of Table~\ref{table_all_miou},
we demonstrate the \texttt{val} and 
\texttt{test} results of WSSS models trained on pseudo masks.
%
Please note that we report only the results on VOC in the main paper, and put the results on MS~COCO in the supplementary materials.
%

Using our AD-CAM achieves $69.0\%$ and $69.4\%$ for PASCAL VOC \texttt{val} and \texttt{test}, respectively. 
Both are clearly
This is significantly better than the top baselines ReCAM and RIB. Note that GETAM adopted DeepLabV2 that was pre-trained on MS COCO, we also apply the pre-trained DeepLabV2 to AD-CAM, it can be seen that AD-CAM is still better than GETAM.

\begin{wraptable}{r}{8cm}
  \vspace{-6mm}
  \centering
  \caption{Compare AD-CAM with baselines in term of consumption time on VOC. The unit time (ut) is 0.7 hours.}
  \label{table_time}
  \setlength{\tabcolsep}{1.2mm}{
  \renewcommand\arraystretch{1}
  \begin{tabular}{lccc}
    \toprule
    Methods & Backbone & Params & time(ut) \\
    \hline
    CAM~\cite{irn}  &  ResNet-50   &  25.6M  & 1.0 \\
    AdvCAM~\cite{anti} & ResNet-50  & 25.6M  &  316.3 \\
    ReCAM~\cite{recam} & ResNet-50  & 25.6M  & 1.9 \\
    \hline
    GETAM~\cite{getam} & ViT-base  & 86M & 23.4 \\
    \hline
    AD-CAM & Conformer-S   &   37.7M  &  2.2 \\
    
    \bottomrule
  \end{tabular}
  }
  \vspace{-4mm}
\end{wraptable}

\noindent
\textbf{Efficiency of AD-CAM.}
Table~\ref{table_time} shows the consumption time of compared methods on VOC. We have the following two observations: 1) Compared with ResNet-50 based methods, AD-CAM uses a slight larger backbone Conformer-S (i.e., 25.6M \texttt{vs} 37.7M), but yields similar computation time compared with ReCAM, i.e., 2.2\textit{ut} \texttt{vs}. And AD-CAM achieves significantly better seeds(63.5 \texttt{vs} 54.8), a small reduction in efficiency is worthable. 2) Compared with ViT- based method, GETAM uses a stronger ViT-base model, which has 86M parameters, but AD-CAM generates better seeds 
This also demonstrates the superiority of AD-CAM.

\subsection{Ablation Study}
\label{sec_ablation}
\begin{wraptable}{r}{6cm}
  \vspace{-10mm}
  \centering
  \caption{Ablation study on VOC (Backbone: Conformer). CAM is the standard CAM extracted from the convolution branch of Conformer. Att-Diffu only means directly performing attention-based diffusion to CAM without attention refinement. AD-CAM$^*$ means without stop-gradient between convolution branch and transformer branch during training. AD-CAM is our final method.}
  \label{table_ablation}
  \setlength{\tabcolsep}{2.2mm}{
  \renewcommand\arraystretch{1.1}
      \begin{tabular}{llccc}
        \toprule
            Methods  & mIoU   & FP & FN  \\ 
            \midrule
             CAM                    & 54.7  & 29.5  & 17.8  \\
             TS-CAM                 & 55.2  & 27.4  & 18.4  \\
             Att-Diffu Only         & 54.6  & 26.4  & 20.3  \\
             AD-CAM$^*$             & 51.7  & 28.8  & 20.8  \\        
             AD-CAM                 & 63.5  & 20.9  & 16.6  \\
        \bottomrule
      \end{tabular}
      }
      \vspace{-4mm}
\end{wraptable}

We perform extensive ablation study on VOC dataset to investigate the effects of different modules of AD-CAM.

\noindent
\textbf{Effect of Co-neighbor AM Refinement.} 
As seen in Table~\ref{table_ablation}, it achieves even worse seeds than standard CAM if we use attention-based CAM diffusion only (54.6 \texttt{vs} 54.7). It can result in more false positive pixels since the initial AM contains many spurious attentions.

\noindent
\textbf{Effect of Stop-gradient.}
In multi-label training, the back propagations between CNN branch and ViT branch are switched off to achieve initial CAM with fewer false positive pixels. To test the effect of this stop-gradient strategy, we also conduct a standard version AD-CAM$^*$. We can see that AD-CAM$^*$ failed to generate comparable CAM, since the crosstalk between two branches will make CNN branch focusing on more global information, thus introduce more false positive pixels.

\subsection{Sensitivity Analysis}
\begin{figure*}[!t]
    \centering
    \includegraphics[width=1.0\linewidth]{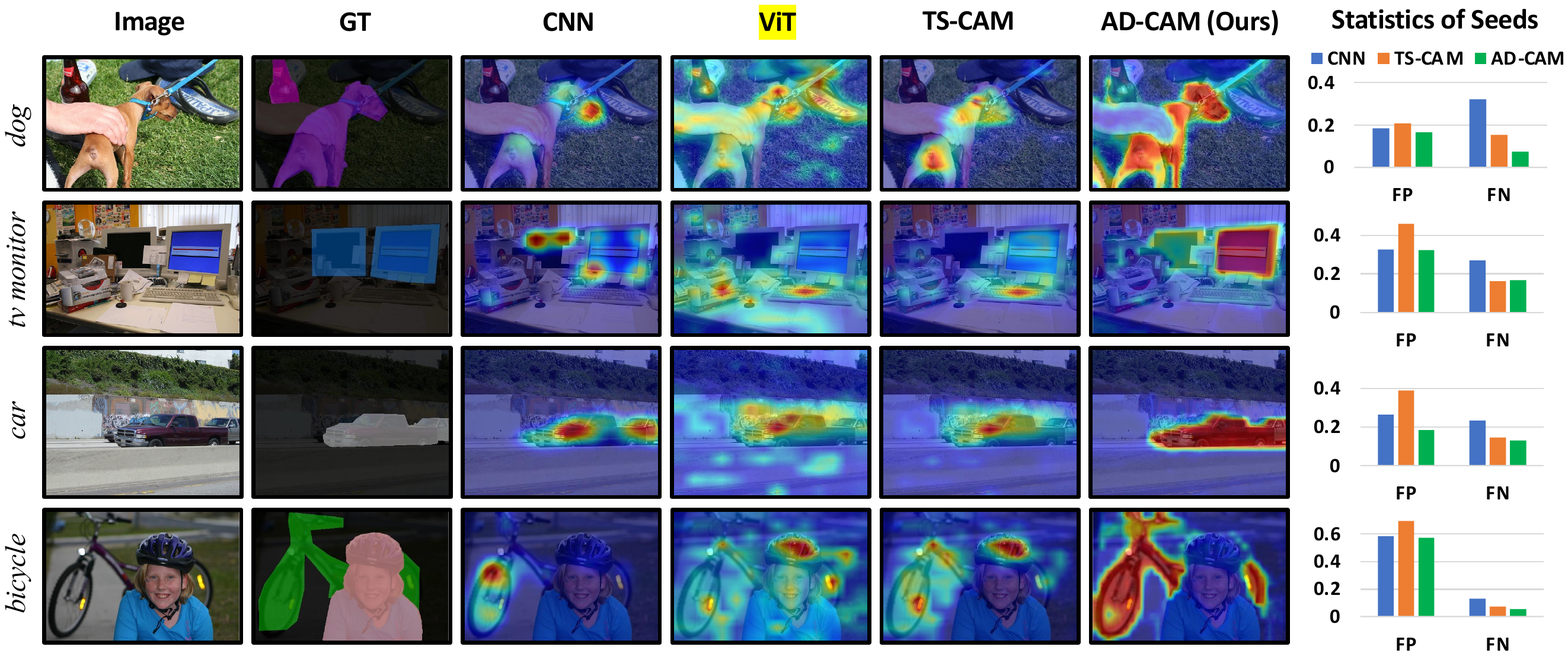}
    \vspace{-4mm}
    \caption{Sensitivity analysis. mIoU, FP and FN are provided. (a) Number of selected top-$k$ attentions based on co-neighbor similarity. (b) Diffusion steps $T$.}
    \label{fig:ablation}
\end{figure*}
\begin{figure*}[!t]
    \centering
    \includegraphics[width=1.0\linewidth]{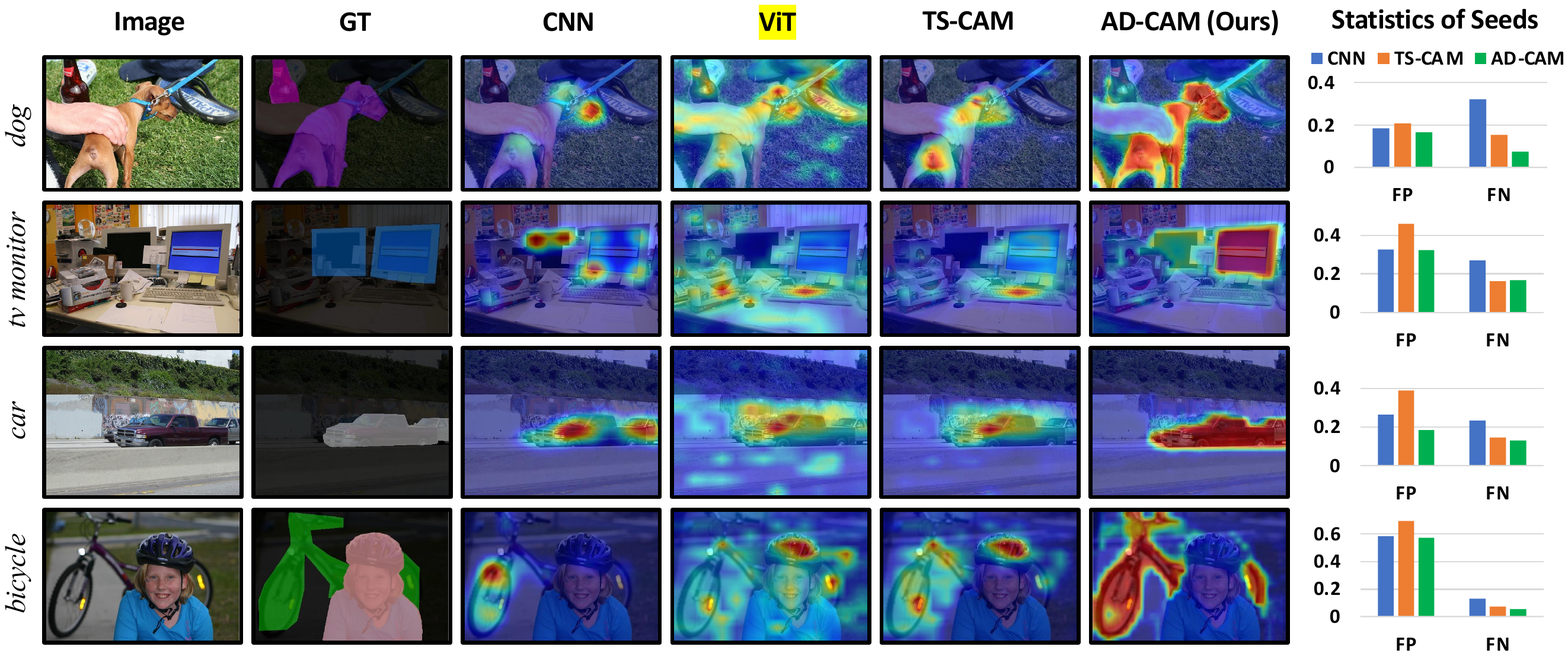}
    \vspace{-4mm}
    \caption{Qualitative Results. CAM heatmaps (i.e., seeds of WSSS) generated by different methods: 1) CNN column is by Conformer conv branch; 2) TS-CAM  on Conformer; 3) AD-CAM* means attn diffusion only; and 4) our AD-CAM column shows our method results based on a double-branch CNN-ViT architecture~\cite{conformer}. 
    For each individual class, we display the statistics of false positive (FP) and false negative (FN) pixels on 0-1 seed masks (generated by applying the optimal threshold of each method on the heatmaps). 
    }
    \label{fig:vis_seed}
    \vspace{-4mm}
\end{figure*}

\begin{figure*}[!t]
    \centering
    \includegraphics[width=1.0\linewidth]{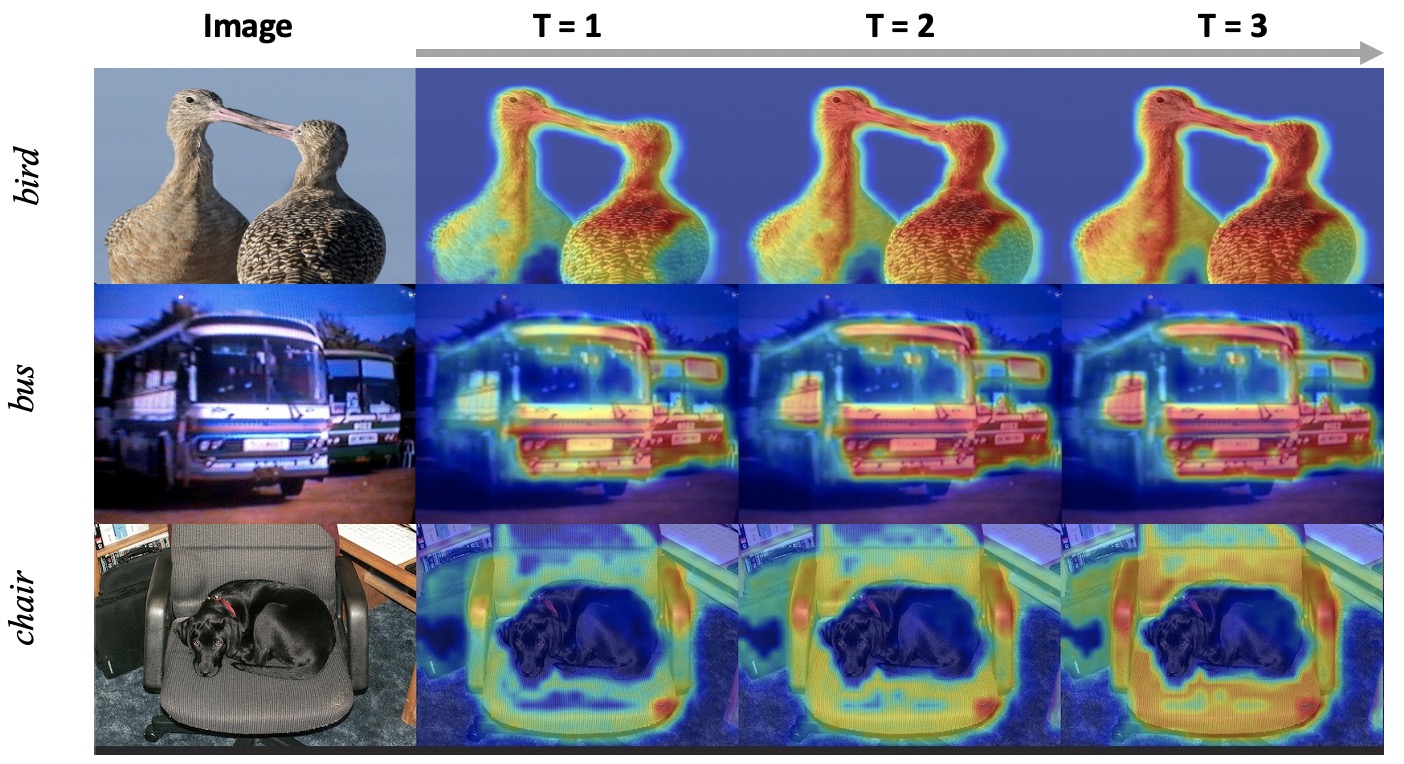}
    \vspace{-4mm}
    \caption{Qualitative Results. CAM heatmaps (i.e., seeds of WSSS) generated by AD-CAM during the attention-based diffusion process. It easy to see the number of false negative pixels significantly drop during the diffusion process.
    }
    \vspace{-4mm}
\end{figure*}

The number of selected top-$k$ attentions based on co-neighbor similarity and diffusion steps $T$ are two important hyper-parameters in our method. We perform sensitivity analysis on VOC dataset by varying $k$ from 20 to 100 and $T$ from 1 to 8. The mIoU, FP and FN are provided in Figure~\ref{fig:ablation}.(1) The mIoU increases rapidly over $k$ with small value, since more and more long-range attentions are included that leads to a significant drop of false negatives(see green line). The performance is relatively stable when $k\geq 50$, so we choose $k$ as 50. (2) The mIoU reaches maximum when $T = 2$, and decreases rapidly when the diffusion process goes further since more false positive pixels are introduced (see orange line).




\section{Conclusion}
Recent works proposed to couple CAM with the long-range attention matrix of visual transformers (ViT) to reduce the false negatives. We find that the attention map of ViT usually contains spurious dependencies, thus leads to a great increase of false positives. To address this issue, we propose AD-CAM to integrate the ViT attention and CAM activation in a probabilistic diffusion way. It first refines the initial attention map by introducing co-neighbor similarity, which take the neighboring information into consideration. Then the attention-based CAM diffusion is achieved by diffusing a pixel’s activation to its neighbors in proportion to the corresponding attentions. Extensive experiments on two challenging WSSS benchmarks PASCAL VOC and MS COCO demonstrates the superiority od AD-CAM.

\clearpage
%
%
\bibliographystyle{splncs04}
\bibliography{egbib}
\end{document}